\documentclass[nonatbib,3p]{elsarticle}
\usepackage{amsmath,amsfonts}
\usepackage{algorithmic}
\usepackage{algorithm}
\usepackage{array}
\usepackage[caption=false,font=normalsize,labelfont=sf,textfont=sf]{subfig}
\usepackage{textcomp}
\usepackage{stfloats}
\usepackage{url}
\usepackage{verbatim}
\usepackage{graphicx}
\usepackage[url=false, isbn=false, sorting=none]{biblatex}
\usepackage{xcolor}
\usepackage{booktabs}
\usepackage[margin=10pt]{caption}
\usepackage[english]{babel}

\AtEveryBibitem{
	\clearfield{note}
}

\begin{document}

\title{On the use of Aggregation Operators to improve Human Identification using Dental Records}

\author[1,2]{Antonio D. Villegas-Yeguas}
\ead{advy99@correo.ugr.es}

\author[2]{Guillermo R-Garc\'ia}
\ead{guiramgar96@gmail.com}

\author[3,5]{Tzipi Kahana}
\ead{kahana.tzipi@gmail.com}

\author[6]{Jorge Pinares Toledo}
\ead{jpinares@odontologia.uchile.cl}

\author[4,5]{Esi Sharon}
\ead{esisharon@gmail.com}

\author[7,2]{Oscar Ib\'a\~nez}
\ead{oscar.ibanez@udc.es}

\author[1]{Oscar Cord\'on}
\ead{ocordon@decsai.ugr.es}

\affiliation[1]{
	organization={Department of Computer Science and Artificial Intelligence, University of Granada},
	city={Granada},
	country={Spain}
}

\affiliation[2]{
	organization={Panacea Cooperative Research S. Coop},
	city={Ponferrada},
	country={Spain}
}

\affiliation[3]{
	organization={Institute of Criminology, The Faculty of Law, Hebrew University of Jerusalem},
	city={Jerusalem},
	country={Israel}
}

\affiliation[4]{
	organization={Department of Prosthodontics, Faculty of Dental Medicine, Hebrew University of Jerusalem},
	city={Jerusalem},
	country={Israel}
}

\affiliation[5]{
	organization={Forensic Odontology Unit, Division of Identification and Forensic Science, Israel Police},
	city={Jerusalem},
	country={Israel}
}

\affiliation[6]{
	organization={Faculty of Dentistry, University of Chile},
	city={Santiago de Chile},
	country={Chile}
}

\affiliation[7]{
	organization={Department of Computer Science and Information Technologies, University of Coru\~na},
	city={A Coru\~na},
	country={Spain}
}

\begin{abstract}
	The comparison of dental records is a standardized technique in forensic dentistry used to speed up the identification of individuals in multiple-comparison scenarios. Specifically, the odontogram comparison is a procedure to compute criteria that will be used to perform a ranking. State-of-the-art automatic methods either make use of simple techniques, without utilizing the full potential of the information obtained from a comparison, or their internal behavior is not known due to the lack of peer-reviewed publications. This work aims to design aggregation mechanisms to automatically compare pairs of dental records that can be understood and validated by experts, improving the current methods. To do so, we introduce different aggregation approaches using the state-of-the-art codification, based on seven different criteria. In particular, we study the performance of i) data-driven lexicographical order-based aggregations, ii) well-known fuzzy logic aggregation methods and iii) machine learning techniques as aggregation mechanisms. To validate our proposals, 215 forensic cases from two different populations have been used. The results obtained show how the use of white-box machine learning techniques as aggregation models (average ranking from 2.02 to 2.21) are able to improve the state-of-the-art (average ranking of 3.91) without compromising the explainability and interpretability of the method.

\end{abstract}


\maketitle

\section{Introduction} \label{introduction}

The last three decades have seen an increase in the number of geopolitical events that claim the lives of numerous victims. Among the main causes of these Disaster Victim Identification (DVI) scenarios are military actions \cite{ashbridge_environmental_2022}, increased natural disasters \cite{ritchie_natural_2024}, and intensified migration crises \cite{mcmahon_death_2021}. These situations of high demand for forensic intervention show an upward trend globally \cite{european_environment_agency_eu_body_or_agency_mapping_2010,us_department_of_justice_namus_2025,evert_unidentified_2011}. Although techniques such as DNA and fingerprints help forensic experts in the task of human identification with a high reliability, these techniques cannot be applied when the remains are taphonomically challenging like in badly burnt and incinerated bodies \cite{cerritelli_interpol_2023} where the soft tissue doesn't survive for friction ridge analysis and DNA sequencing cannot be performed \cite{forrest_forensic_2019, cerritelli_interpol_2023-1}. Depending on the technology available to the DVI professionals, as well as the accessibility of \textit{ante-mortem} (AM) data, dental identification have proven to be highly effective in some DVI scenarios \cite{perrier_swiss_2006, kirsch_tsunami_2007}.

Teeth contain the hardest tissues in the human body and are highly resistant to decomposition, high temperatures and other extreme environmental conditions, including traumatic, physical, and chemical factors. The adult dentition normally consist of 32 teeth, each with a complex anatomy characterized by at least five surfaces (mesial, occlusal, distal, facial, lingual) and one or more roots. Every anatomical feature has its own morphology and may be affected by various lesions or restorations (pathological lesions, esthetic or metallic restorations, tooth anomalies, etc.). The vast number of possible combinations of these characteristics creates significant variability, which is highly valuable for personal identification. For all these reasons forensic odontology is considered one of the primary identification techniques by Interpol, at the same level at fingerprint analysis or DNA profile comparisons \cite{forrest_forensic_2019,pittayapat_forensic_2012,sweet_oc_interpol_2010, cerritelli_interpol_2023}.

Forensic dental identification relies on the comparison of dental records of a missing person with those of a deceased individual. Such records may include written treatment notes, radiographs, casts, photographs, and CT scans. The process involves comparing AM and \textit{post-mortem} (PM) dental characteristics, which are documented as codes on odontograms --- that is, schematic records of each tooth's condition. Identification is established by evaluating the concordance between the AM and PM odontograms, while accounting for logical changes that may occur in the teeth of the missing person over time. Specifically, the odontogram comparison is a procedure to compute criteria that will be used to perform a ranking.  This procedure is usually addressed using automatic data science and artificial intelligence methods, as the numbers of comparisons grow exponentially with the number of cases, and it is not feasible to manually perform all those comparisons. 

Last July 2024, the AI Act was released by the European Commission. As part of the first AI regulation in Europe, the High-Level Expert Group on AI developed the Assessment List for Trustworthy Artificial Intelligence (ALTAI) \cite{directorate-general_for_communications_networks_content_and_technology_european_commission_ethics_2019, directorate-general_for_communications_networks_content_and_technology_european_commission_assessment_2020}

This tool defines recommendations and requirements for AI systems according to a risk level classification. In particular, automatic biometric methods for human identification are classified as a "high rist (acceptable)". This type of systems have string requirements in terms of: human agency and oversight, technical robustness and safety, privacy and data governance, transparency diversity, non-discrimination and fairness, environmental and societal well-being and accountability. Despite dental records comparison being a major pillar in human identification \cite{pittayapat_forensic_2012, krishan_dental_2015, ata-ali_forensic_2014, devadiga_whats_2014, waleed_importance_2015}, the majority of current methods fail on achieving ALTAI requirements mainly due to the lack of validation and access to the inner workings of the algorithms.

The few existing approaches considering trustworthiness by design \cite{mcgivney_computer-assisted_2001, adams_computerized_2016} have important technical limitations, severely affecting accuracy and robustness. They all use a lexicographical order to generate the ranking. This implies that they do not use all the information available, as the criteria are ordered by importance, in most cases will only be used to resolve ties in the process of generating a ranking.

For all these reasons, in this study we propose explainable aggregation methods to obtain a score to improve the accuracy of these rankings. Several approaches are proposed and tested. The different proposals are validated using data from forensic scenarios from two different populations, considering a 5-fold cross validation (CV) data split to train and validate the machine learning models, and a test split to compare all the proposals.

First, in section II we review how odontograms work, their relevance, how they are being used in forensic odontology and the main problems faced in order to evolve to more trustworthy solutions. Then, in section III we introduce our proposal to improve the state-of-the-art method that involves odontogram comparison. In section IV we apply the current method and our proposals to data from forensic scenarios to evaluate their performance. Finally, in section V we discuss the results and their applicability in real-world scenarios by forensic experts.

\section{Background: Use of odontograms in forensic odontology and explainable artificial intelligence} \label{background}
\noindent
An odontogram is a form containing coded information about the different characteristics and pathologies of the teeth of an individual. These types of records can be very useful, as they equate the AM and PM records making the comparison process easier. The AM records are obtained from previous medical records of missing persons, which can be stored in a wide variety of forms such as written descriptions, panoramic radiographs or full mouth radiographs, cone beam computed tomography (CBCTs), computed tomography (CT), and intraoral 3D scans. Meanwhile, PM records are obtained from the human remains through CT scans, X-ray images, and written records through direct observation of the cadaver's dentition. The use of odontograms solves the problem of working with multiple data types, when performing an identification the expert can use information in the same format to compare the AM and PM profiles.

These forms, when used within an automatic system, are extremely useful to shortlist candidates in large databases, reducing the number of comparisons needed to achieve a positive identification. For this reason they have been widely used in DVI scenarios \cite{perrier_swiss_2006,beauthier_mass_2009,kirsch_tsunami_2007}.

Human identification by odontogram comparison involves working with a set of AM records and a single PM record and vice versa and analyzing the coincidences and discrepancies between each AM-PM comparison. The goal is to find out which AM or PM record have more coincidences and less non-explainable discrepancies. In order to perform this task we need to figure out some details of how odontograms works, how the information is codified in the form, and how the comparison between them is performed.

\subsection{Codification system}

To represent the dentition state on a form it is first necessary to establish which characteristics are to be represented in the record, i.e. the dental coding system. There are many different proposals of codification systems in the field. They mainly differ in the granularity level considered, from one unique code per tooth to multiple codes per each tooth surface and root \cite{chomdej_intelligent_2006, hashimoto_coding_1984, lorton_computer-assisted_1989, adams_computerized_2016}. Most of these proposals share codes for the most common status, such as a \textit{virgin}, \textit{missing} either AM or PM, \textit{filled}, \textit{implant} or \textit{rotated} tooth, and the granularity refers to the specification of these statuses, going into detail of the affected surface of the tooth, or the material of the interventions.

In the absence of a single international standard, Interpol codification proposal (currently composed of 41 codes, although previous versions of the codification system consisted of more than 130 codes) represents the codification with a wider consensus \cite{franco_feasibility_2013, franco_comparing_2019-1}. Even so, some other codifications are still in use, such as the codification proposed by Adams and Aschheim in \cite{adams_computerized_2016} as a simplification of the WinID3 \cite{al-amad_evaluation_2007} codification system with only 7 codes.

The main reason why there are different codification systems with different granularity levels is that we are facing a very difficult problem with a tradeoff between discriminative power and errors due to the difficulty to label accurately using a large dataset of codes. When using a very detailed system we can find ambiguous scenarios, where more than one status can be assigned to a tooth or to a part of a tooth, leading to errors in the data collection and/or problems when making the comparison (in both cases, greater expertise is required by the practitioner to label the odontogram and compare the AM and PM data). In contrast, when a simple codification is used, the available codes may group different conditions together, preventing the system from properly individualizing subjects.

\subsection{Odontogram comparison and rankings}
\label{odontogram_comparison_rankings}

The odontogram comparison process must be developed keeping in mind that the pair of AM and PM records can or can not belong to the same person. Hence, this process tells the expert whether a pair of AM-PM records is plausible, i.e. whether the changes in the dentition observed in the PM record with respect to the AM record are consistent or can be explained by the course of time, such as the dental interventions. This reason leads to using a ranking system when comparing odontograms. Instead of returning the most plausible record the system will rank all the records available in descending order of plausibility.

The rankings are not based on single criteria, but in some of them measuring different aspects. The state-of-the-art systems use the following ones \cite{dailey_computer-assisted_1987, mcgivney_computer-assisted_2001, adams_computerized_2016}:

\begin{itemize}
  \item Match
  \item Possible match
  \item Mismatch
\end{itemize}

When comparing two records, each system implements a lookup table to check whether: i) the state of a tooth is the same in the AM and PM record (a match), ii) the status in the AM record is different from the status in the PM record but the difference can be explained (possible match, e.g. the tooth is virgin in the AM record but has a filling in the PM record), or iii) the status in both records is different and cannot be explained (mismatch, e.g. the tooth has a filling in the AM record but is virgin in the PM record).

Software systems have been developed over the years to perform this task automatically. One of the first popular and widely used systems for odontogram comparison was CAPMI \cite{lorton_computer-assisted_1989}, developed in the 1980s by the US Army Institute of Dental Research. In a first approach this system considered if a tooth was present or not, and which surface has been restored for each tooth. Then, with the development of new additional systems, it was expanded to include information about secondary characteristics to add details about the materials used in restorative treatments, aiming to perform more detailed comparisons. Due to the fact that CAPMI was developed by the US Army, the scenarios in which it has been widely used are in military conflicts \cite{dailey_computer-assisted_1987} or US nationals departments \cite{haglund_national_1993}.

Another extended software is WinID, a successor to CAPMI developed in 1992 \cite{mcgivney_computer-assisted_2001}. WinID uses the same primary codes as CAPMI, four primary codes to describe whether the tooth is \textit{unerupted}, \textit{virgin}, \textit{missing} or \textit{fractured}, another code for no information on the tooth, one code for the crown, and one code per tooth surface (mesial, occlusal, distal, facial, lingual). The main addition of WinID was the use of secondary codes, optional modifiers to represent extra information such as restoration materials, in scenarios with multiple subjects with similar records. This added a more detailed comparison only when needed, keeping data collection simple and avoiding errors related to this step.

In 1997 a project to mimic the Interpol paperwork for DVI in the Victorian Institute of Forensic Medicine, Australia, led to the development of DAVID (Disaster And Victim IDentification), mostly used by Australian authorities with good results \cite{clement_new_2006}. This proposal uses a matrix to assign a numeric value to each combination of codes, obtaining a score for each comparison.

Later, in 2004, researchers from Thailand proposed IDIS \cite{chomdej_intelligent_2006}, a software to perform odontogram comparison using 29 codes to describe the tooth state and based on the guidelines and standards for DVI of the American Board of Forensic Odontology and Interpol.

Last but not least, probably the most widely used software, specially outside the US, is KMD PlassData DVI, supported by Interpol \cite{sweet_oc_interpol_2010}. Initially developed by Plass Data (a Danish small and medium enterprise, later acquired by the multinational KMD) its initial objective was to replicate all of Interpol's protocols and bureaucratic processes for human identification. This software allows the comparison of odontograms based on the matches and misses between AM and PM records, computing a similarity score. However, the method used to compute this score is not public. KMD PlassData DVI has been used in multiple scenarios, such as the Thailand tsunami from 2004, the Air France AF447 plane crash in the Atlantic Ocean, and the Germanwings A320 plane crash in France.

\subsection{State-of-the-art limitations}

In addition to the advantages mentioned above, the use of odontogram comparison has certain disadvantages. As we discussed before, one of the main problems with odontograms is choosing a good codification system, one that shows enough individualization power to be useful in human identification but simple enough to avoid errors in data collection and to keep the codification understandable.

Other issues we observe in the state-of-the-art methods are closely related to the key requirements of ALTAI. Regarding the human agency and oversight, the proposals discussed do not provide the oversight mechanisms to allow the experts to make informed decisions. This is mainly caused by the lack of three other requirements: transparency, technical robustness and accountability. In this regard, those systems do not explain their inner inference process, have not been validated, and do not have audits to verify their results, so it is also unknown whether they are sufficiently robust. For example, there is not any public statistical study showing the number of rankings positions that an expert should check to find the positive case with a certain coverage. Finally, concerning to diversity, non-discrimination and fairness, the datasets used to create these systems, and therefore possible biases derived from the data, are unknown.

Despite these problems being well known in the literature, there are very few proposals to solve them. One of the few examples in the literature is the Adams and Aschheim's proposal \cite{adams_computerized_2016} to simplify the WinID3 system by reducing the number of codes to seven. In their publication, they clearly describe how to generate the ranking and which characteristics to use, furthermore the system is validated and benchmarked with WinID3 using both forensic and synthetic data.

\subsection{The Simplified Coding System for odontograms}

The Simplified Coding System (SCS) \cite{adams_computerized_2016} use seven codes applied to the whole tooth. This codification has proven to have a significant individualization ability despite using a simple detailed description. As said, the use of a simpler system avoids errors in data collection due to mistakes in annotation or ambiguity, especially when using 2D radiographic images where is not possible to distinguish details with such precision when obtaining an odontogram.

The codes used in this codification are:

\begin{itemize}
  \item \textbf{V}: The tooth is virgin or unerupted.
  \item \textbf{F}: The tooth is present and has a filling.
  \item \textbf{S}: The tooth is present and has a special treatment (e.g. a root canal).
  \item \textbf{X}: The tooth was missing AM.
  \item \textbf{I}: The tooth has been replaced with an implant.
  \item \textbf{P}: The tooth is or was present, but a treatment is not visible for some reason (e.g., not fully visible in X-ray, a fracture)
  \item \textbf{N}: No information available at all.
\end{itemize}

With these codes we can use tables \ref{tab:scsPostmortem} and \ref{tab:scsAntemortem} to compare a pair of odontograms and obtain the number of matches, possibles matches and mismatches.

\begin{table}[H]
\centering
\begin{tabular}{@{}ccll@{}}
\cmidrule(l){2-4}
\multicolumn{1}{l}{}         & \multicolumn{3}{c}{Comparison result by AM code}                   \\ \midrule
\multicolumn{1}{c|}{PM Code} & Match & \multicolumn{1}{c}{Mismatch} & \multicolumn{1}{c}{Possible} \\ \midrule
\multicolumn{1}{c|}{V}       & V     & F, S, X, I                   & P                            \\
\multicolumn{1}{c|}{F}       & F     & S, X, I                      & V, P                         \\
\multicolumn{1}{c|}{S}       & S     & X, I                         & F, V, P                      \\
\multicolumn{1}{c|}{X}       & X     &                              & V, F, S, I, P                \\
\multicolumn{1}{c|}{I}       & I     &                              & V, F, S, X, P                \\
\multicolumn{1}{c|}{P}       &       & X, I                         & V, F, S, P                   \\
\multicolumn{1}{c|}{N}       &       &                              &                              \\ \bottomrule
\end{tabular}
\caption{Classification matrix for each possible PM-AM comparison in SCS.}
\label{tab:scsPostmortem}
\end{table}

\begin{table}[H]
\centering
\begin{tabular}{@{}ccll@{}}
\cmidrule(l){2-4}
\multicolumn{1}{l}{}         & \multicolumn{3}{c}{Comparison result by PM code}                    \\ \midrule
\multicolumn{1}{c|}{AM Code} & Match & \multicolumn{1}{c}{Mismatch} & \multicolumn{1}{c}{Possible} \\ \midrule
\multicolumn{1}{c|}{V}       & V     &                              & F, S, X, I, P                \\
\multicolumn{1}{c|}{F}       & F     & V                            & S, X, I, P                   \\
\multicolumn{1}{c|}{S}       & S     & V, F                         & X, I, P                      \\
\multicolumn{1}{c|}{X}       & X     & V, F, S, P                   & I                            \\
\multicolumn{1}{c|}{I}       & I     & V, F, S, P                   & X                            \\
\multicolumn{1}{c|}{P}       &       &                              & V, F, S, X, I, P             \\
\multicolumn{1}{c|}{N}       &       &                              &                              \\ \bottomrule
\end{tabular}
\caption{Classification matrix for each possible AM-PM comparison in SCS.}
\label{tab:scsAntemortem}
\end{table}

Note that, when we do not have information about at all about a tooth, the result of a comparison is unknown; it is neither a match, a mismatch, nor a possible match.

\subsection{Adams and Aschheim ranking generation\label{AdamsProposal}}

Using the SCS to represent the dental characteristics in the records, the authors propose to take into consideration the following criteria in order to generate the ranking:

\begin{itemize}
    \item Number of Matches
    \item Number of Possible matches
    \item Number of Mismatches
    \item Percentage of Matches
    \item Percentage of Possible matches
    \item Percentage of Mismatches
    \item Exact group
\end{itemize}

Where exact group is a discretization of the percentage of matches, transforming this percentage into the following levels:

\begin{itemize}
    \item Group 0: 0\% of matches.
    \item Group 1: 1-44\% of matches.
    \item Group 5: 45-49\% of matches.
    \item Group 5.5: 50-54\% of matches.
    \item Group 6: 55-59\% of matches.
    \item Group 6.5: 60-64\% of matches.
    \item Group 7: 65-69\% of matches.
    \item Group 7.5: 70-74\% of matches.
    \item Group 8: 75-79\% of matches.
    \item Group 8.5: 80-84\% of matches.
    \item Group 9: 85-90\% of matches.
    \item Group 10: 90-100\% of matches.
\end{itemize}

After trying different combinations they propose to shortlist the ranking by the exact group, and in case of tie, use the number of matches to break the tie. For the case where both the exact group and the number of matches are equal, they use the lowest percentage of mismatches to break the tie. Finally, if there is a tie while using these three criteria, they use the percentage of possible matches to break the tie. This type of ordering method is known as lexicographical ordering in the field of fuzzy sets and systems \cite{farhadinia_hesitant_2016}.

In this work we used the SCS codification to represent the dental records in our dataset, and extended the method proposed by Adams and Aschheim in different ways.

\section{Proposal} \label{proposal}
\noindent

Our study has two main objectives, the first one is to obtain reliable and robust rankings, and the second one is that the obtained method is interpretable. This helps forensic experts understand how the model obtains the scores in order to rank the comparisons and gain scientific knowledge by understanding how the system works. For this reason we focused mainly on explainable methods such as fuzzy aggregators and interpretable machine learning models. In addition, we used a more complex black-box machine learning model to numerically benchmark and discuss the main differences that can be found regarding the trade-off between explainability and performance \cite{rudin_stop_2019}.

Specifically, in this work we propose three different approaches: i) using a data-driven approach to design an aggregation mechanism similar to the state-of-the-art method, ii) using advanced aggregation operators, iii) using machine learning models as aggregation mechanisms to learn the most accurate criteria combinations.

\subsection{Data-driven lexicographical order-based aggregation mechanism}

We believe that by ranking first by the lower percentage of mismatches followed by the higher number of matches, the system will be able to avoid scenarios where a comparison with a high number of matches, but a few mismatches would be better positioned than a comparison with no mismatches, many matches and a few possible matches. For example, following the proposal of Adams and Aschheim, a comparison with 30 matches and 2 mismatches is better positioned than a comparison with 28 matches and 4 possible matches. With this solution we expect to find an improvement of the original proposal, maintaining the explicability and the simplicity of the ranking generation.

As we can test the system with different combinations, our following proposal is to use a data-driven approach to explore all possible combinations of criteria used in the Adams and Aschheim's proposal.

\subsection{Exploring fuzzy aggregation operators}

As said, there is a need to aggregate multiple criteria in order to create a score-based ranking. In our case, we want to aggregate all the different criteria obtained from the comparison of two odontograms to obtain a value that represents the similarity between these two cases. Aggregation operators are basic tools to achieve this goal \cite{beliakov_aggregation_2007} and have been widely used in forensic scenarios with good results \cite{campomanes-alvarez_experimental_2016, anderson_estimation_2010}.

In this work we propose to use two different types of aggregation operators, ordered-based \cite{yager_ordered_1993, xu_fuzzy_2009} and fuzzy integral-based \cite{sugeno_theory_1974, choquet_theory_1954} aggregation operators. The latter type of aggregators are known for being one of the most powerful and flexible operators, since they use fuzzy measures to represent the relevance of each criterion.

\subsubsection{Ordered Weighted Averaging (OWA)}

The OWA operator was proposed by Yager \cite{yager_ordered_1993} to solve multi-criteria decision-making problems. The particularity of this proposal with a classical weighted average is that the coefficients are not associated directly with the criteria, they are associated to a certain ordered position. This idea can be very useful in our case as with this proposal we will not associate weights to a certain criterion, we propose a method where in each comparison between two odontograms we can give more weight to the most present criteria.

Given a set of weights $W = [w_{1}, \dots, w_{n}]$ and a set of inputs $A = [a_{1}, \dots, a_{n}]$ the proposed aggregation function is:

\begin{equation}
\label{owaFunction}
  F(a_{1}, \dots, a_{n}) = \sum_{i=1}^{n}{w_{i}b_{i}}
\end{equation}

\noindent where $b_{i}$ is the $i^{th}$ largest element of $A$. By choosing a different weight vector $W$ we can implement different aggregation operators, allowing us to be more flexible when testing this approach.


\subsubsection{Ordered Weighted Harmonic Mean (OWHM)}

The ordered fuzzy harmonic mean operator was proposed by Xu in 2009 \cite{xu_fuzzy_2009}. It allows us to apply classic harmonic mean to fuzzy values, developing certain operators such as fuzzy weighted harmonic mean, or the one considered in this case, the ordered fuzzy harmonic mean.

Given a set of inputs $A = \{a_{1}, \dots, a_{n}\}$, the Harmonic Mean is defined as:

\begin{equation}
\label{harmonicMean}
    HM(a_{1}, \dots, a_{n}) = \frac{n}{\sum_{i=1}^{n}{ \frac{1}{a_{i}}}}
\end{equation}

Xu proposes to introduce the Weighted Harmonic Mean aggregator by adding a set of weights $W = \{w_{1}, \dots, w_{n}\}$, this aggregator performs the reciprocal of each value $a_{n}$ using the weight $w_{n}$, obtaining the following aggregator:

\begin{equation}
\label{weightedHarmonicMean}
    WHM(a_{1}, \dots, a_{n}) = \frac{n}{\sum_{i=1}^{n}{ \frac{w_{i}}{a_{i}}}}
\end{equation}

\noindent with $w_{j} \ge 0$.

Following the OWA approach, Xu also proposes a variation in which the set of inputs are ordered, and then the weights are applied to the ordered items, instead of associating a certain weight to a certain position of the input, obtaining the fuzzy OWHM:

\begin{equation}
\label{orderedWeightedHarmonicMean}
    OWHM(a_{1}, \dots, a_{n}) = \frac{n}{\sum_{i=1}^{n}{ \frac{w_{i}}{b_{i}}}}
\end{equation}

\noindent where $b_{i}$ is the $i^{th}$ largest element of $A$.

\subsubsection{Choquet fuzzy integral}

The Choquet integral \cite{choquet_theory_1954} is a widely used aggregator in fuzzy logic, commonly used in multi-criteria decision problems. Let $X = \{x_{1}, \dots, x_{n}\}$ be a finite set of attributes, and let $F$ be the family of all subsets of $X$, a fuzzy measure $\mu$ on $X$ is a set function $\mu : F \rightarrow [0, 1]$, which meets the following conditions:

\begin{enumerate}
    \item $\mu( \emptyset ) = 0$, $\mu(X) = 1$
    \item Given $A \subseteq B$, then $\mu(A) \le \mu(B)$ for all $A, B \in F$
\end{enumerate}

Then, we define a Choquet integral of a function $f : X \rightarrow [0, 1]$ with respect to the fuzzy measure $\mu$ as:

\begin{equation}
\label{choquetIntegral}
    C_{\mu}(f) = \sum_{i=1}^{n}{ [f(x_{i}) - f(x_{i - 1})}] \mu(A_{i})
\end{equation}

\noindent where $x_{0} = 0$ by convention and $f(x_{i})$ is a permutation so that $0 \le f(x_i) \le 1$, $A_{i} = x_i, \dots, x_n$.

\subsubsection{Sugeno fuzzy integral}

Another commonly used fuzzy integral is the Sugeno integral \cite{sugeno_theory_1974}. We define a Sugeno integral of a function $f: X \rightarrow [0, 1]$ with respect to the fuzzy measure $\mu$ as

\begin{equation}
\label{sugenoIntegral}
    S_{\mu}(f) = \vee_{i = 0}^{n} [ f(x_{i}) \wedge \mu(A_i) ]
\end{equation}

\noindent with the same notation as in \eqref{choquetIntegral}, and where $\vee$ is a t-conorm and $\wedge$ is a t-norm, generally the maximum and minimum respectively.

When using the Choquet and Sugeno integrals, first we need to define the measure $\mu$ that we will use in the equations \eqref{choquetIntegral} and \eqref{sugenoIntegral}. In our case we launched the experiments with two measures, the cardinality and a Sugeno $\lambda$-measure \cite{sugeno_fuzzy_1993}.

A fuzzy measure $\mu_{\lambda}$ is a Sugeno $\lambda$-measure if there exists a $\lambda > -1$ such that for all $A, B \in F$:

\begin{equation}
\label{lambdaMeasureUnion}
    \mu_{\lambda}(A \cup B) = \mu_{\lambda}(A) + \mu_{\lambda}(B) + \lambda \mu_{\lambda}(A)\mu_{\lambda}(B)
\end{equation}

Also note that for a Sugeno $\lambda$-measure, and because of the boundary condition, the following expression holds:

\begin{equation}
\label{lambdaSum}
    \prod_{i}^{n}(1 + \lambda \mu_{\lambda}^{i}) = 1 + \lambda
\end{equation}

The values $\mu_{\lambda}^{1}, \dots, \mu_{\lambda}^{n}$ are called the fuzzy densities, that in our case can be interpreted as the importance of the individual methods. To obtain a value for $\lambda$ we will use the equation \eqref{lambdaSum}, using the normalized average ranking obtained by considering only the variable $i$ to generate the rank as the value of $\mu_{\lambda}^{i}$. In this way, we are assigning the relevance of each feature based on its discriminatory power. After that, we can use \eqref{lambdaMeasureUnion} to compute the measure for any input set.

\subsection{Machine learning models as aggregators}
\noindent

In this approach we will use machine learning model as aggregation mechanisms. Instead of designing or using already established aggregation operators, we will train machine learning models to learn the best way to aggregate the considered criteria.

\subsubsection{Linear Regression (LR)} For each criterion obtained in the comparison of two odontograms we will try to find a weight to compute the score of each comparison by:

\begin{equation}
\label{linearRegression}
    LR(X) = \beta + \sum_{i=1}^{n}{w_i * x_i}
\end{equation}

\noindent given an input $X = [x_{1}, \dots, x_{n}]$ and a set of weights $W = [w_{1}, \dots, w_{n}]$, where $\beta$ as the independent term of the regression.

In our case, as we used the seven criteria proposed by Adams and Aschheim \cite{adams_computerized_2016} (see Section \ref{AdamsProposal}), we need to find the weights for seven input variables and one for the independent term. LR models are often fitted using the least squares approach, as the training dataset is labelled. But in our case our goal was to minimize the position of the true comparison in a ranking, and the model is used to compute the score to rank the comparisons, which is unknown a priori. For this reason we cannot use least squares, so we used a genetic algorithm in order to find the weights of the regression.

\subsubsection{Symbolic Regression (SR)}

SR is a type of regression that, instead of fixing the regression equation and then learning the weights of each term of the regression, it searches the space of mathematical expressions to also obtain the structure of the regression. A key advantage of his approach is the model's impartiality as it is not affected by the modeler's bias. Many approaches typically use evolutionary algorithms as search and optimization techniques in order to explore the large solution space offered by symbolic regression. Genetic programming (GP) is a type of evolutionary algorithm proposed by Koza \cite{koza_genetic_1992} that follows all the fundamentals principles of genetic algorithms, using a tree-based grammar to represent individuals. This algorithm represents the mathematical expression as a tree, where the leaf nodes represent constants and input variables, and the branch nodes encode mathematical operators. With this representation, the crossover and mutation operators are applied to both the tree nodes and subtrees.

SR and GP are especially useful for problems where transparency and interpretability are crucial \cite{mei_explainable_2023}, as ours. With this approach the aggregation mechanism itself will be a mathematical expression, which can be interpreted by the expert using it, giving the possibility to validate and understand its inner workings.

\subsubsection{Multilayer Perceptron (MLP)}

In the last decades neural networks have proven an outstanding performance in many different data science and machine learning problems \cite{lecun_deep_2015}. Our last proposal is to use a MLP to obtain a score to create a ranking. As our loss function is not derivable, we cannot use the commonly used back-propagation algorithm in order to train the neural network, so we will use a genetic algorithm in order to find the weights of the network.

We propose a basic neural network architecture with an input layer of seven neurons, one for each input criteria, three hidden layers of 32 neurons each one, and finally an output layer of one neuron that will return the score of one comparison.

Although this proposal does not allow us to obtain an interpretable model by the expert \cite{rudin_stop_2019,barredo_arrieta_explainable_2020,longo_explainable_2024}, we believe that the versatility and good results of these models can be used as a non-interpretable benchmark for comparing the different proposed methods.

\section{Experiments and analysis of results} \label{experiments}
\noindent

To analyze the performance of our proposal, including a comparison with the state-of-the-art method proposed by Adams and Aschheim \cite{adams_computerized_2016}, several experiments have been carried out. In this section we will first describe the available dataset and the experimental setup designed. Then, we will go into the experiments performed for each proposed approach, as well as a comparison between all the different approaches considered.

\subsection{Experimental design}

In order to obtain the necessary data, various dentists and specialists in the field of human identification were contacted, providing us with data from both Israel and Chile. The data from Israel (Helsinky HMO-02450-25) consist of a total of 2850 radiographic images, from 256 cases, both orthopantomographies and periapical radiographs, as well as dental records described according to Interpol codification (133 codes), collected between 2006 and 2016 from real forensic cases. The data from Chile consisted of a total of 559 orthopantomographies, collected between 2004 and 2018 by the CIMEX dental center. In cases where the odontogram was already filled in by the specialists, we checked that the condition described therein coincided with that of the X-ray whenever possible, as well as adapting the coding used by the specialist to the coding used in our experiments, SCS (7 codes). In the remaining cases, the odontogram was filled based on the dental conditions observed in the X-ray by two of the authors, Dr. Kahana and Mr. R-Garc\'ia. After reviewing these data and filtering the cases for which both AM and PM information were available, a total of 215 paired cases were selected, providing a total of 430 dental charts. In both datasets, the PM data was collected at least three years after the AM case. 

As the proposed machine learning-based aggregation mechanisms need a training step, we made a split of data into training, validation, and test. Since we consider two different datasets, the split between training, validation and test sets was done for each of them, to avoid possible over-fitting due to different data provenance. For each dataset, we performed a 20\% holdout to obtain the test partition, and then a 5-fold CV on the remaining 80\% of cases: create five partitions with the same number of cases in each, and train the model five times using the different combinations of four partitions and leaving the fifth partition for validation. This methodology allowed us to perform a model selection without biasing the best model to perform better in the test split.

While we could use the whole dataset for the experiments testing the first two approaches, i.e. on data-driven lexicographical order and fuzzy aggregations (as these methods do not require a training step), we also used the same test split to make results comparable with the machine learning models.

\begin{table}[H]
\centering
\begin{tabular}{@{}lccc@{}}
\toprule
                                & \multicolumn{1}{l}{Israel} & \multicolumn{1}{l}{Chile} & \multicolumn{1}{l}{\textbf{Total}} \\ \midrule
\multicolumn{1}{l|}{Training}      & 83                        & \multicolumn{1}{c|}{56}    & 139                                \\
\multicolumn{1}{l|}{Validation} & 20                        & \multicolumn{1}{c|}{14}    & 34                                 \\
\multicolumn{1}{l|}{Test}       & 25                        & \multicolumn{1}{c|}{17}    & 42                                 \\ \midrule
    \textbf{Total}                  & 128                       & 87                         & 215                                \\ \bottomrule
\end{tabular}

\caption{Data split by number of cases. Training and validation refers to the numbers of cases in one fold.}
\label{tab:datasplit}
\end{table}

To evaluate the different approaches we used the ranking of each case. When comparing one PM case versus $N$ AM cases, we will obtain a ranking, as explained on Section \ref{odontogram_comparison_rankings}. In that ranking of $N$ elements, we refer to the correct comparison position for the position where the AM case is the same as the PM case. This value is between 1 and $N$. The closer this position is to 1, the better, since that tell us that the system is able to identify the correct AM case and position it at the top of the ranking. As we will obtain $N$ rankings, one per case as we perform a comparison of $N$ AM cases versus $N$ PM cases, we can compute multiple statistics for each method. Specifically we will compute the average correct comparison position (in average, in which position of a ranking is the correct comparison), the maximum position (from all the rankings, the higher correct comparison position), the quartiles (Q1, Q2 and Q3, indicate the ranking position covering 25\%, 50\% and 75\% of the correct comparison positions, respectively), and the 95th and 99th percentile (indicate the ranking position covering 95\% and 99\% of the correct comparison positions, respectively). For clarity, we will refer to the value of a ranking as the position of the correct comparison, i.e, a ranking of 3 means that the position of the correct comparison is 3.

These statistics allowed us to study each proposal from different perspectives, as a good proposal is not only one in which the mean ranking is low (in average, we can find the positive comparison in the first positions), but also one where the maximum ranking is not very high (for the worst case we do not need to review every case), even if the last one implies a worse mean ranking. We also used the cumulative match curve (CMC) in order to visualize the ranking capabilities of each system \cite{li_handbook_2011}. In the CMC plots we visualized the percentage of cases that have a correct comparison position lower than or equal to the ranking position represented on the X-axis. For this reason, the length of the X-axis is the number of cases used for the comparison that generated the ranking, 42 in the case of the test dataset.

Another thing to keep in mind is that when comparing the statistics between the training, validation and test sets, the number of cases within each dataset is different, so we cannot make a direct comparison of these values. For this reason, in tables comparing train and validation results, the values were normalized dividing by the number of cases in each dataset.

Regarding the operators and hyperparameters for the methods trained with genetics algorithms, we used the following values:

\begin{itemize}
    \item Population size: 100 chromosomes
    \item Initial population: Genes with random values in the range $[-50, 50]$
    \item Tournament type: Roulette wheel selection
    \item Tournament size: 50 individuals
    \item Encoding scheme: Real numbers
    \item Operators (only for the symbolic regression): Addition, subtraction, division and multiplication
    \item Crossover operator:
        \begin{itemize}
            \item LR: BLX-$\alpha$ \cite{eshelman_real-coded_1993}, $\alpha = 0.5$
            \item GP: Swap subtree
            \item MLP: BLX-$\alpha$, $\alpha = 0.5$
        \end{itemize}
    \item Mutation operator:
        \begin{itemize}
            \item LR: Random mutation
            \item GP: Random point mutation and subtree mutation
            \item MLP: Random mutation
        \end{itemize}
    \item Crossover probability: 70\%
    \item Mutation probability: 5\%
    \item Generations: 1000
    \item Fitness function: Minimize average ranking
\end{itemize}

These hyperparameters have been chosen empirically after a study of which hyperparameters allow all the machine learning models to converge. This way all the proposed models had the same configuration and none of them was penalized for having a slower convergence. Regarding the value to be minimized during the optimization process, we have run two experiments, one minimizing the average ranking and the other minimizing the maximum ranking, obtaining similar results. Hence, for the sake of clarity, we will only show and discuss the former.

\subsection{Data-driven lexicographical order-based aggregation}

For these experiments we have evaluated all the combinations of criteria in order to find out which lexicographical order-based aggregation mechanism obtained the best results. We found out that the best design for our dataset was the following one:

\begin{enumerate}
    \item{\% Mismatch}
    \item{\% Match}
    \item{\# Mismatch}
    \item{\# Match}
    \item{Exact group}
    \item{\% Possible}
    \item{\# Possible}
\end{enumerate}

\begin{figure}[H]
\centering
\includegraphics[width=3.5in]{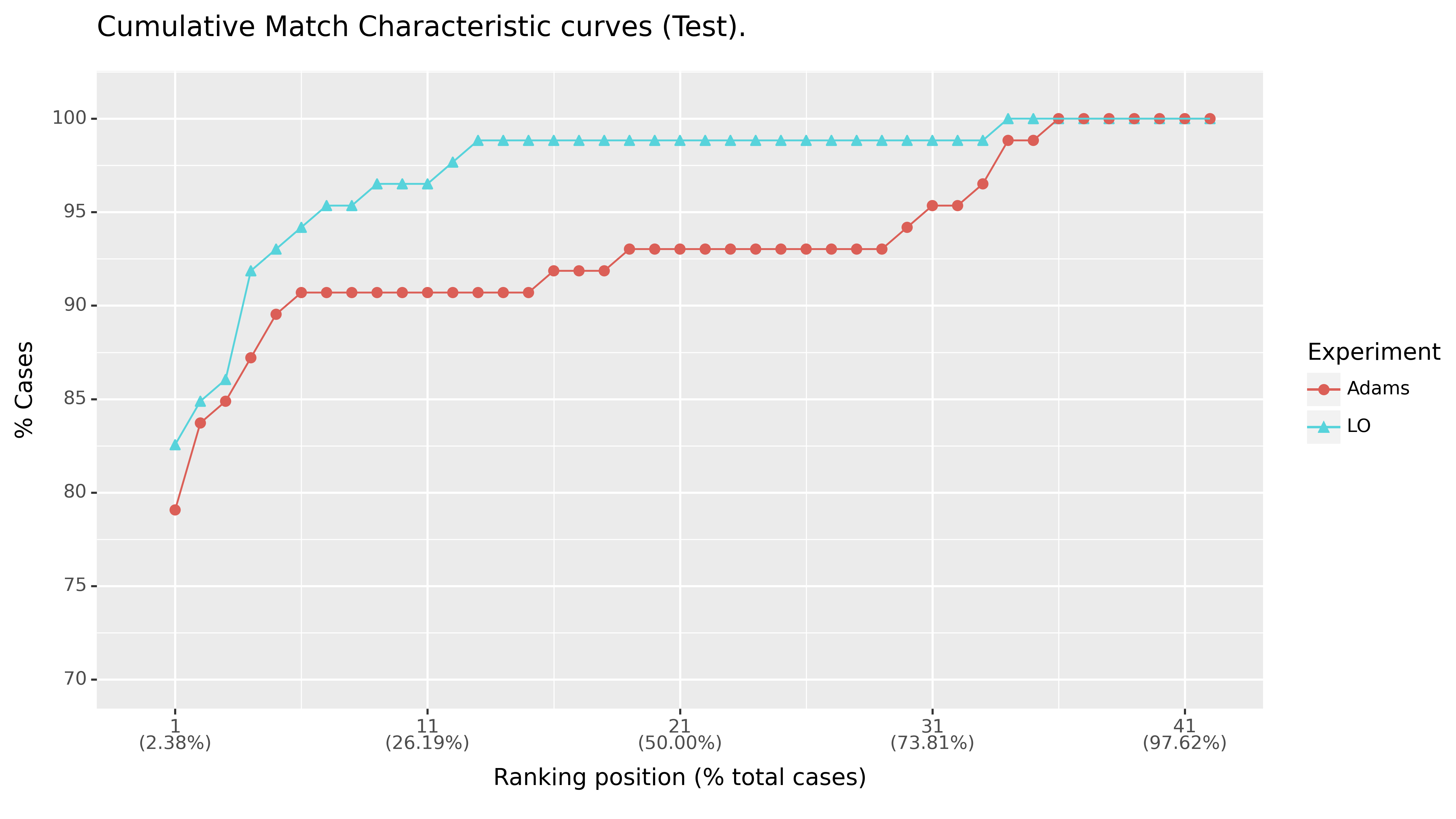}
	\caption{CMC plot comparing the results of our LO aggregation mechanism and Adams and Aschheim's proposal in the test dataset. }
\label{cmcExpertKnowledge}
\end{figure}

As expected with the lexicographical order-based (LO) aggregation mechanism, the best combinations start with the same two first criteria. If we take a look at the statistics (Table \ref{tab:expertKnowledgeStats}) and the CMC plot (Fig. \ref{cmcExpertKnowledge}), we can clearly see an improvement when using the proposed solution with respect to Adams and Aschheim's proposal:

\begin{table}[H]
\centering
\begin{tabular}{@{}lccllllc@{}}
\toprule
                                                 & \multicolumn{1}{l}{Average} & \multicolumn{1}{l}{Q1} & Q2 & Q3 & P95 & P99 & \multicolumn{1}{l}{Max} \\ \midrule
\multicolumn{1}{l|}{AA} & 3.91                        & 1                      & 1  & 1  & 30  & 34  & 36                      \\
\multicolumn{1}{l|}{LO}                        & 2.14                        & 1                      & 1  & 1  & 6   & 16  & 34                      \\ \bottomrule
\end{tabular}
\caption{Comparison of ranking performance between Adams and Aschheim's proposal (AA) and our LO aggregation mechanism.}
\label{tab:expertKnowledgeStats}
\end{table}

With this simple proposal we nearly double the accuracy of the average ranking, dramatically improving the 95th and 99th percentile, and slightly lowered the maximum ranking. This is a clear example of how using a data-driven approach we can design a better aggregation mechanism for odontogram comparison while keeping it simple.

According to all the experiments carried out, the worst combinations of criteria are those that used either the number or the percentage of possible matches in first place. This can be explained by the great uncertainty of these criteria, since they are very common and do not provide much information.


\subsection{Fuzzy aggregation operators}

\subsubsection{Ordered Weighted Averaging}

For the OWA operator, we tried different sets of weights to test different scenarios:

\begin{itemize}
    \item Maximum: For each comparison between two odontograms, the criteria with the maximum value is chosen. In this case, the set of weights $W$ will have $w_1 = 1$ and $w_i = 0$ for $i \ne 1$.
    \item Minimum: For each comparison between two odontograms, the criteria with the minimum value is chosen. We do this in order to, when generating the ranking with all the comparisons, choose the comparison that has the best minimum support of all its variables. The set of weights $W$ will have $w_n = 1$ and $w_i = 0$ for $i \ne n$.
    \item Average: The final score is the average of all the criteria. The set of weights $W$ will have $w_i = \frac{1}{n}$ for all $i \in [1, n]$.
    \item Linear: In this case the criteria with less support will have less weight in the final decision, and those with a strong support will have more decision power. The set of weights $W$ will have $w_i = i$ for all $i \in [1, n]$.
\end{itemize}

With all those different sets of weights we obtained the following results in the test set:

\begin{table}[H]
\centering
\resizebox{0.49\textwidth}{!}{%
\begin{tabular}{@{}lccllllc@{}}
\toprule
                                                 & \multicolumn{1}{l}{Average} & \multicolumn{1}{l}{Q1} & Q2 & Q3 & P95 & P99 & \multicolumn{1}{l}{Max} \\ \midrule
	\multicolumn{1}{l|}{AA} & 3.91                        & 1                      & 1  & 1  & 30  & 34  & 36                      \\
\multicolumn{1}{l|}{OWA-Maximum}                        & 5.9                        & 1                      & 4  & 7  & 18   & 25  & 35                      \\
\multicolumn{1}{l|}{OWA-Minimum}                        & 4.99                        & 1                      & 1  & 3  & 27   & 37  & 39                      \\
\multicolumn{1}{l|}{OWA-Average}                        & 3.27                        & 1                      & 1  & 1  & 15   & 26 & 36                      \\
\multicolumn{1}{l|}{OWA-Linear}                        & 2.86                        & 1                      & 1  & 1  & 11   & 22  & 36                      \\ \bottomrule
\end{tabular}
}
	\caption{Comparison of ranking performance between Adams and Aschheim's proposal (AA) and our approach using the OWA operator with different weights.}
\label{tab:OWAStats}
\end{table}

Table \ref{tab:OWAStats} shows how using basic operators such as the minimum or the maximum yields worse results, while averaging the criteria, or using linear weights, giving more importance to the criteria that most supports the decision, improves the results. If we look at the CMC curves (Fig. \ref{cmcOWA}) we can see this behavior in detail:

\begin{figure}[H]
\centering
\includegraphics[width=3.5in]{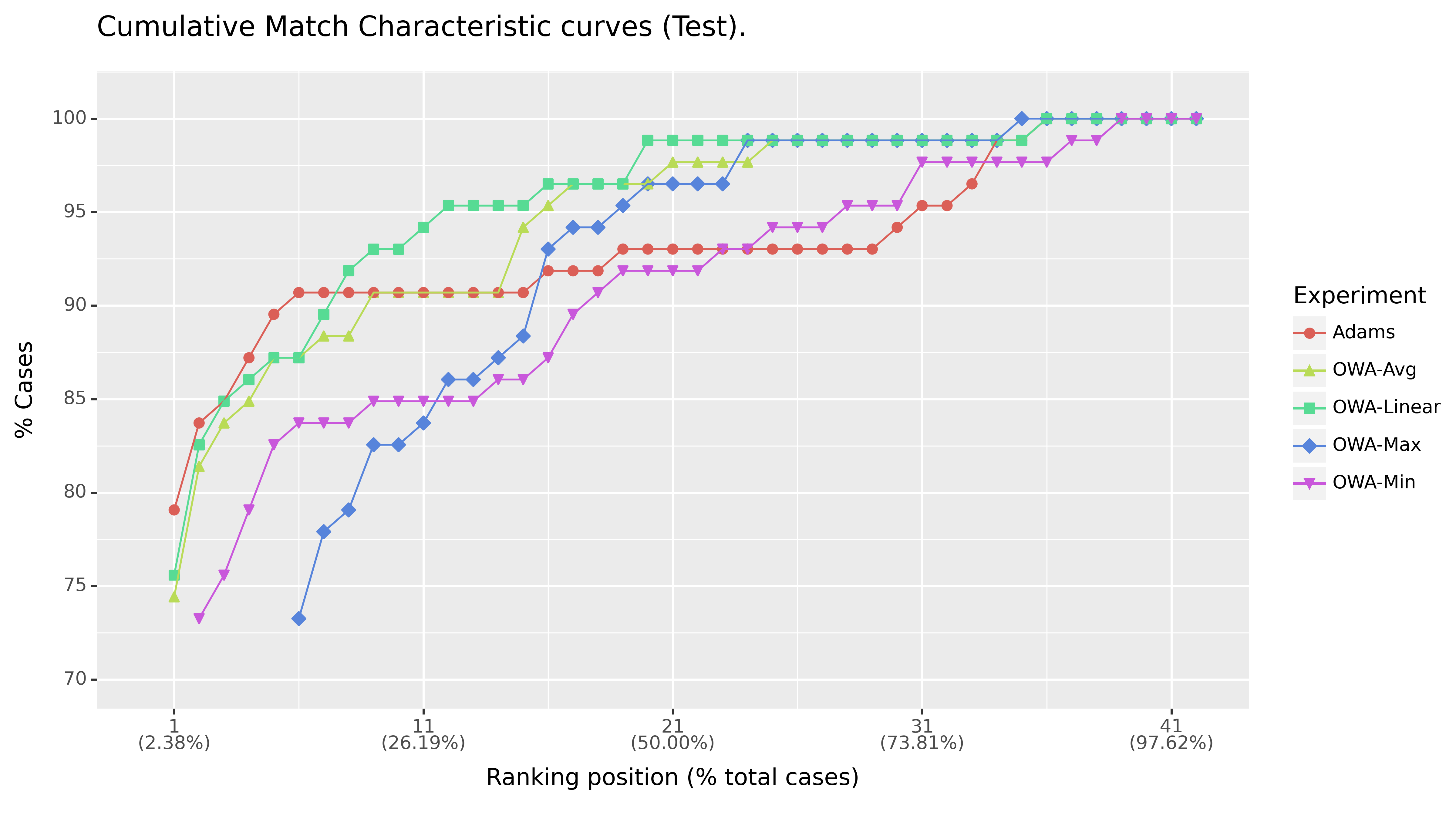}
\caption{CMC plot comparing the results of our proposal of the OWA operator with Adams and Aschheim's proposal in the test dataset. }
\label{cmcOWA}
\end{figure}

Although the state-of-the-art proposal performs better in the first positions of the ranking, the OWA proposal with the linear weights and the proposal with the average weights greatly improves the ranking after the 15\% of ranking positions, reducing the number of positions needed to cover more than the 95\% of cases. The average and the linear OWA aggregations, while obtaining worse results than Adams and Aschheim's proposal for the starting positions, maintain a reasonable coverage of cases when looking only the first positions of the rankings, while the minimum and maximum weights proposal obtains much worse results. Regarding the latter method, the maximum aggregator is able to outperform the state-of-the-art proposal when looking at the 35\% of positions in all rankings, as shown in figure \ref{cmcOWA}.

\subsubsection{Ordered Weighted Harmonic Mean}

For this aggregator we used the same set of weights except for the maximum and minimum one, as we would get the same results, but instead of just getting the maximum or the minimum, we would get $\frac{n}{b_i}$ where $b_i$ is the maximum or the minimum of the input set. We obtained the following results:

\begin{table}[H]
\centering
\resizebox{0.49\textwidth}{!}{%
\begin{tabular}{@{}lccllllc@{}}
\toprule
                                                 & \multicolumn{1}{l}{Average} & \multicolumn{1}{l}{Q1} & Q2 & Q3 & P95 & P99 & \multicolumn{1}{l}{Max} \\ \midrule
\multicolumn{1}{l|}{AA} & 3.91                        & 1                      & 1  & 1  & 30  & 34  & 36                      \\
\multicolumn{1}{l|}{OWHM-Average}                        & 6.51                        & 1                      & 2  & 6  & 28   & 36  & 39                      \\
\multicolumn{1}{l|}{OWHM-Linear}                        & 6.02                        & 1                      & 2  & 5  & 28   & 36  & 38                      \\ \bottomrule
\end{tabular}
}
	\caption{Comparison of ranking performance between Adams and Aschheim's proposal (AA) and our approach using the OWHM aggregator with different weights.}
\label{tab:OWHMStats}
\end{table}

These results are much worse than the Adams and Aschheim's proposal, both in average and in maximum ranking. We can clearly see that this operator does not perform well in the CMC curve (Fig. \ref{cmcOWHM_0_100}):

\begin{figure}[H]
\centering
\includegraphics[width=3.5in]{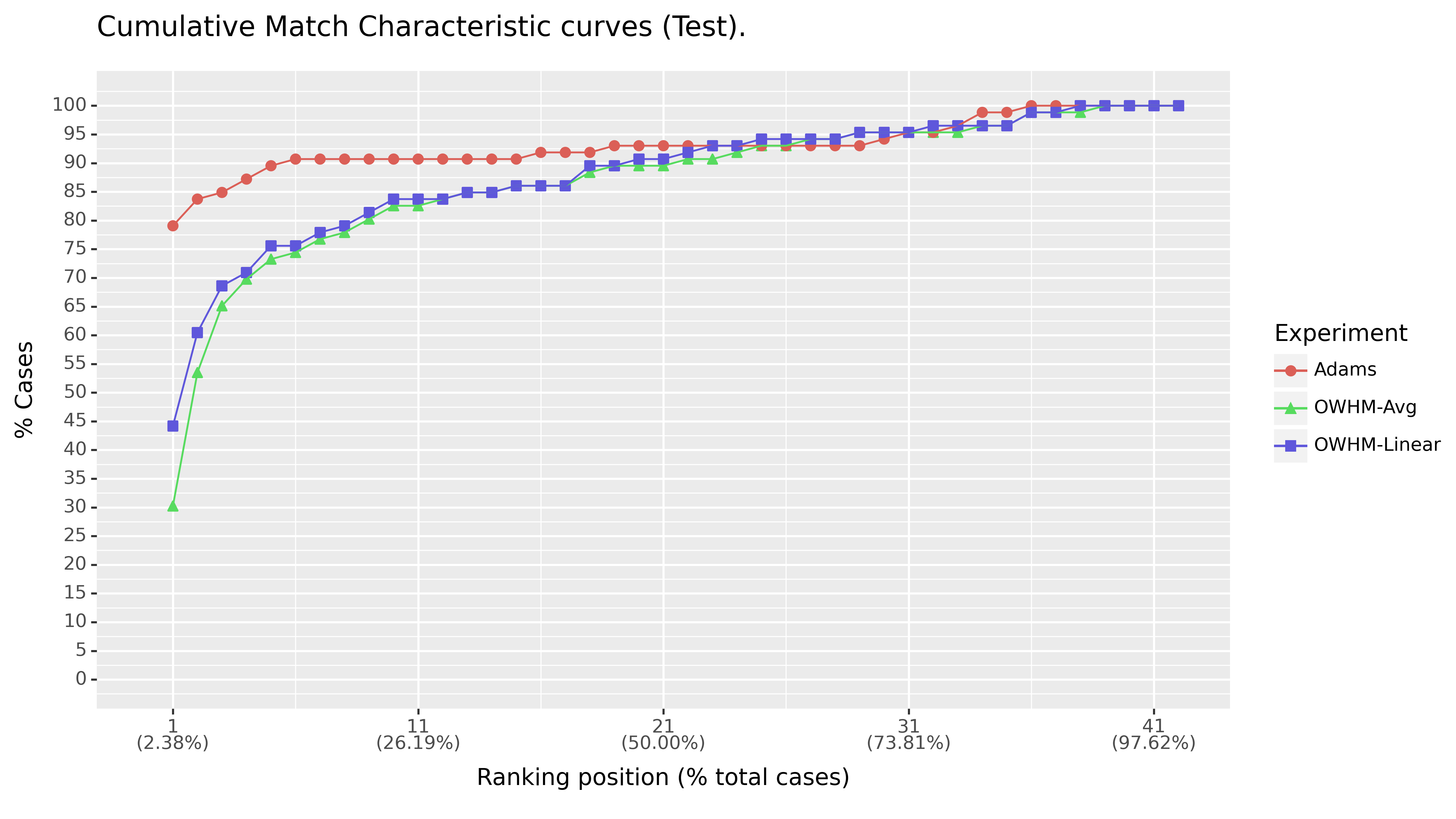}
\caption{CMC plot comparing the results of our proposal of the OWHM aggregator with Adams and Aschheim's proposal in the test dataset. Notice that differently to previous CMC plots, the Y axis range is 0-100\%.}
\label{cmcOWHM_0_100}
\end{figure}

In this case, using a more conservative aggregator has penalized the final ranking. This may be due to the fact that in this problem we have a large uncertainty on the part of some criteria, which means that if we are conservative when aggregating their information, instead of dissipating that uncertainty with the use of other criteria, we would impose it.

\subsubsection{Choquet integral}

In these experiments we obtained the following results:

\begin{table}[H]
\centering
\resizebox{0.49\textwidth}{!}{%
\begin{tabular}{@{}lccllllc@{}}
\toprule
                                                 & \multicolumn{1}{l}{Average} & \multicolumn{1}{l}{Q1} & Q2 & Q3 & P95 & P99 & \multicolumn{1}{l}{Max} \\ \midrule
\multicolumn{1}{l|}{AA} & 3.91                        & 1                      & 1  & 1  & 30  & 34  & 36                      \\
\multicolumn{1}{l|}{\begin{tabular}[c]{@{}l@{}}Choquet integral:\\ \hspace{0.4cm} $\mu$: Cardinality\end{tabular}}                        & 3.26                        & 1                      & 1  & 1  & 15   & 26  & 36                      \\
\multicolumn{1}{l|}{\begin{tabular}[c]{@{}l@{}}Choquet integral:\\ \hspace{0.4cm} $\mu$: $\lambda$\end{tabular}}                        & 3.14                        & 1                      & 1  & 1  & 16   & 33  & 35                      \\ \bottomrule
\end{tabular}
}
	\caption{Comparison of ranking performance between Adams and Aschheim's proposal (AA) and the Choquet integral.}
\label{tab:choquetStats}
\end{table}

Table \ref{tab:choquetStats} shows that the Choquet integral slightly improved the state-of-the-art proposal when using either of the two measures, obtaining better results if we use the Sugeno-$\lambda$ measure, as it includes information on the importance of each criterion. With the CMC curve (Fig. \ref{cmcChoquet}) we can visualize this behavior in more detail:

\begin{figure}[H]
\centering
\includegraphics[width=3.5in]{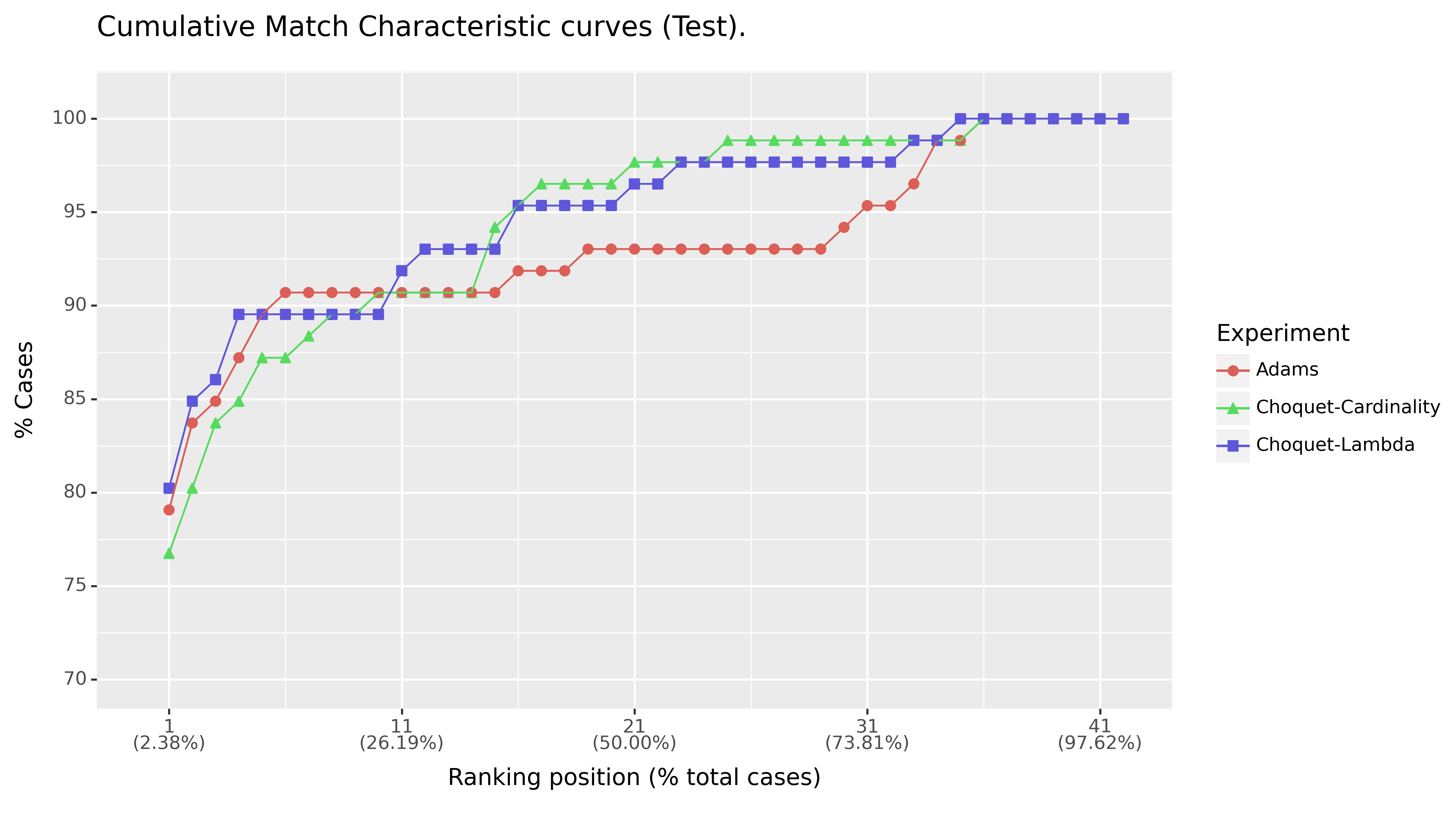}
\caption{CMC plot comparing the results of our proposal of the Choquet integral aggregator with Adams and Aschheim's proposal in the test dataset. }
\label{cmcChoquet}
\end{figure}

At first, when looking the first ranking positions, the Choquet integral with the Sugeno $\lambda$-measure are very similar and cover almost the same percentage of cases, but when we look at lower ranking positions our proposals need less positions to cover the same percentage of cases.

\subsubsection{Sugeno integral}

For the Sugeno integral we used the same measures as for the Choquet integral, obtaining the following results:

\begin{table}[H]
\centering
\resizebox{0.49\textwidth}{!}{%
\begin{tabular}{@{}lccllllc@{}}
\toprule
                                                 & \multicolumn{1}{l}{Average} & \multicolumn{1}{l}{Q1} & Q2 & Q3 & P95 & P99 & \multicolumn{1}{l}{Max} \\ \midrule
\multicolumn{1}{l|}{AA} & 3.91                        & 1                      & 1  & 1  & 30  & 34  & 36                      \\
\multicolumn{1}{l|}{\begin{tabular}[c]{@{}l@{}}Sugeno integral:\\ \hspace{0.4cm} $\mu$: Cardinality\end{tabular}}                        & 6.06                        & 1                      & 1  & 6  & 27   & 34  & 40                      \\
\multicolumn{1}{l|}{\begin{tabular}[c]{@{}l@{}}Sugeno integral:\\ \hspace{0.4cm} $\mu$: $\lambda$\end{tabular}}                        & 5.28                        & 1                      & 3  & 5  & 18   & 25  & 35                      \\ \bottomrule
\end{tabular}
}
	\caption{Comparison of ranking performance between Adams and Aschheim's proposal (AA) and the Sugeno integral.}
\label{tab:sugenoStats}
\end{table}

When using the Sugeno integral the overall results are clearly worse than the Adams and Aschheim's proposal. If we look at the quartiles in Table \ref{tab:sugenoStats} we can see that the main difference between the Sugeno integral with the $\lambda$-measure and the Adams and Aschheim's proposal is that the former performs much worse at the first positions on the ranking, covering much fewer cases, while greatly improving the results at the final ranking positions, as we can see in figure \ref{cmcSugeno}:

\begin{figure}[H]
\centering
\includegraphics[width=3.5in]{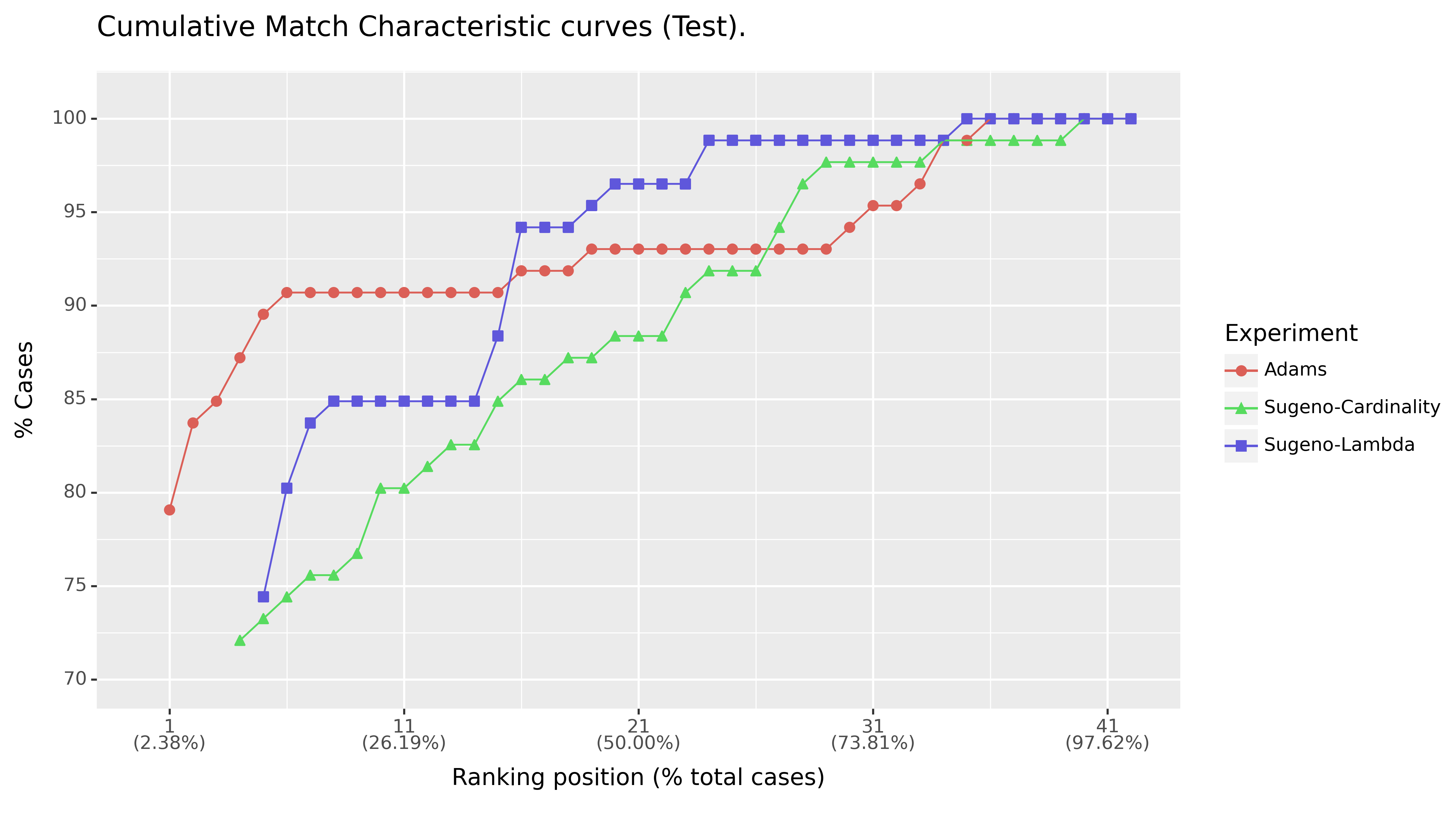}
\caption{CMC plot comparing the results of our proposal of the Sugeno integral aggregator with Adams and Aschheim's proposal in the test dataset. }
\label{cmcSugeno}
\end{figure}

A detail to take into account is how the Sugeno integral proposal with the Sugeno-$\lambda$ measure is able to improve the results for the most difficult comparisons, obtaining better results from the 95th percentile onwards, although it obtains a worse average ranking due to its poor performance in the first positions of the ranking.

\subsection{Machine learning models}

\subsubsection{Linear regression}

We obtained the following average rankings using the 5-folds CV:

\begin{table}[H]
\centering
\resizebox{0.49\textwidth}{!}{%
\begin{tabular}{@{}ccccc@{}}
\toprule
    Fold                      & \multicolumn{2}{c}{Training} & \multicolumn{2}{c}{Validation} \\ \cmidrule(l){2-5}
                          & Average     & Maximum     & Average            & Maximum   \\ \midrule
\multicolumn{1}{c|}{1}    & 0,0061      & 0,3478      & 0,0041             & 0,0857    \\
\multicolumn{1}{c|}{2}    & 0,0058      & 0,3043      & 0,0055             & 0,1429    \\
\multicolumn{1}{c|}{3}    & 0,0041      & 0,1449      & 0,0139             & 0,3714    \\
\multicolumn{1}{c|}{4}    & 0,0065      & 0,3381      & \textbf{0,0009}    & \textbf{0,0294}    \\
\multicolumn{1}{c|}{5}    & 0,0074      & 0,3165      & 0,0013             & \textbf{0,0294}    \\ \midrule
Average of folds          & 0,0060      & 0,2904      & 0,0051             & 0,1318   \\ \bottomrule
\end{tabular}
}
\caption{Normalized results of the average and maximum ranking obtained using a genetic algorithm to learn the weights of a linear regression. In bold the best validation results.}
\label{tab:foldsLR}
\end{table}

As Table \ref{tab:foldsLR} shows, all the folds obtain similar results in average ranking regardless considering the training or validation dataset. We can see big differences in the maximum ranking, as there are folds where the validation result is much worse, like fold number three, and folds with much better results like number four and five.

To select our final model we used the validation values of the 5-fold CV, so in this case the best model is the obtained with the fold 4, which corresponds to the following aggregation mechanism expression:

\begin{equation}\begin{split}
\label{linearRegressionScoreFunction}
    LR(X) = 0.67 - 0.2891x_{1} + 0.4594x_{2} \\
          - 0.8470x_{3} + 0.0214x_{4} - 0.1328x_{5} \\
          + 0.0544x_{6} - 0.9323x_{7}
\end{split}\end{equation}

\noindent where $x_1$ is the exact group, $x_2$ the number of matches, $x_3$ the number of mismatches, $x_4$ the percentage of matches, $x_5$ the percentage of possible matches, $x_6$ the number of possibles matches, and $x_7$ the percentage of mismatches.

Interpreting this result from an expert's point of view, the weights for both $x_3$ and $x_7$ are both negative and penalize the final score, as having numerous mismatches is a bad aspect when comparing odontograms. Besides, the weight for $x_2$, the number of matches, clearly the most important positive weight, as it is positive to contain as many matches as possible. For the weight associated to $x_5$ we can see that it penalizes the final score, but not as much as the number of mismatches, since having a possible match is a source of uncertainty when comparing odontograms. Regarding the weights for $x_4$ and $x_6$, they have almost no contribution to the final score as they are close to zero, probably because the information encoded in these criteria is the same as in $x_2$ and $x_5$ respectively, thus being redundant. Finally, an interesting issue about the weight for $x_1$ is that, although it represents the percentage of matches, it has a negative value. We think that this is because this value moves from a low value to a high value (from group 1 to group 5) with a small variation of the number of matches, so it is not very descriptive in a comparison.

\begin{table}[H]
\centering
\resizebox{0.49\textwidth}{!}{%
\begin{tabular}{@{}lccllllc@{}}
\toprule
                                                 & \multicolumn{1}{l}{Average} & \multicolumn{1}{l}{Q1} & Q2 & Q3 & P95 & P99 & \multicolumn{1}{l}{Max} \\ \midrule
\multicolumn{1}{l|}{AA} & 3.91                        & 1                      & 1  & 1  & 30  & 34  & 36                      \\
\multicolumn{1}{l|}{LR}                        & 2.21                        & 1                      & 1  & 1  & 6   & 23  & 24                      \\
\multicolumn{1}{l|}{SR}                        & 2.02                        & 1                      & 1  & 1  & 4   & 23  & 30                      \\
\multicolumn{1}{l|}{MLP}                        & 1.94                        & 1                      & 1  & 1  & 4   & 18  & 28                      \\ \bottomrule
\end{tabular}
}
	\caption{Comparison of ranking performance between Adams and Aschheim's proposal (AA) and our machine learning-based aggregation mechanisms.}
\label{tab:MLStats}
\end{table}

\begin{figure}[H]
\centering
\includegraphics[width=3.5in]{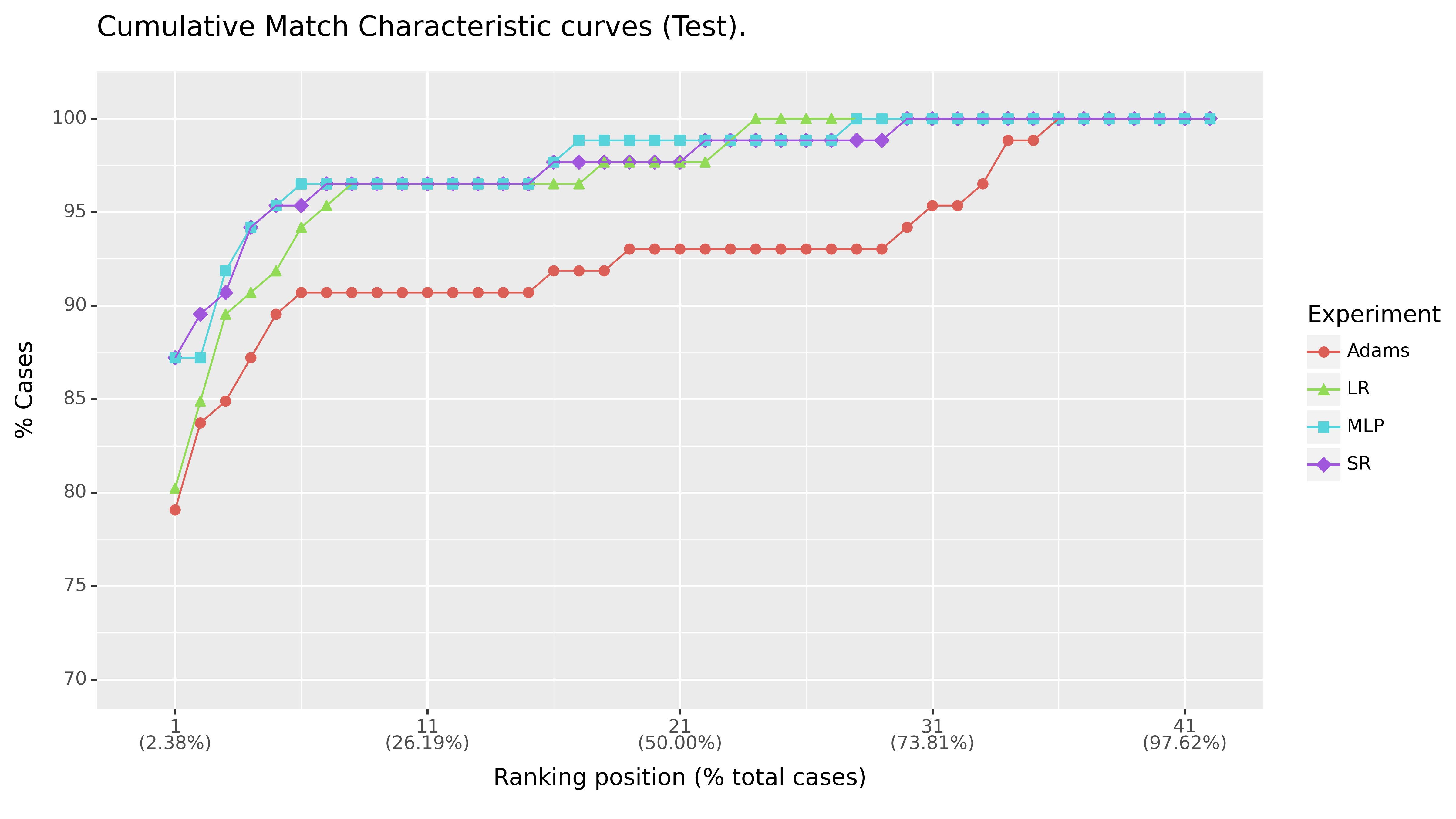}
\caption{CMC plot comparing the results of our proposal of machine learning-based aggregation mechanisms as score generator with Adams and Aschheim's proposal in the test dataset. }
\label{cmcML}
\end{figure}

In Table \ref{tab:MLStats} and Figure \ref{cmcML} we can see that, as expected, we improved the Adams and Aschheim's method in both average and maximum ranking by aggregating information from all the criteria. These improvements are not only seen in some cases, but in general, even for the more difficult cases.

\subsubsection{Symbolic regression}

When launching the 5-folds CV we obtain the results in Table \ref{tab:foldsSR}:

\begin{table}[H]
\centering
\resizebox{0.49\textwidth}{!}{%
\begin{tabular}{@{}ccccc@{}}
\toprule
Fold                   & \multicolumn{2}{c}{Training} & \multicolumn{2}{c}{Validation} \\ \cmidrule(l){2-5}
                       & Average     & Maximum     & Average            & Maximum   \\ \midrule
\multicolumn{1}{c|}{1} & 0,0059      & 0,3188      & 0,0069             & 0,1429    \\
\multicolumn{1}{c|}{2} & 0,0047      & 0,2754      & 0,0101             & 0,3429    \\
\multicolumn{1}{c|}{3} & 0,0039      & 0,1594      & 0,0184             & 0,9714    \\
\multicolumn{1}{c|}{4} & 0,0048      & 0,2662      & \textbf{0,0009}    & \textbf{0,0294}    \\
\multicolumn{1}{c|}{5} & 0,0049      & 0,2374      & 0,0013             & \textbf{0,0294}    \\ \midrule
Average of folds       & 0,0049      & 0,2514      & 0,0075             & 0,3032    \\ \bottomrule
\end{tabular}
}
\caption{Normalized results of the average and maximum ranking obtained using a genetic algorithm to learn a symbolic regression-based aggregation mechanism. In bold the best validation results.}
\label{tab:foldsSR}
\end{table}

In this case we can see that the model has a similar average ranking performance in the training and validation datasets. As in linear regression, in the folds 2 and 3 we can see that the maximum ranking is much worse than in others folds, maybe in these folds a really hard case is present. Regarding the best model, the one with the best validation error is the fold number 4, which obtains the following equation:

\begin{equation}
\label{symbolicRegressionScoreFunction}
    SR(X) = \frac{\frac{x_1 - x_6}{x5}}{x_6}
\end{equation}

The interpretation of this result is even simpler than for the linear regression. Here the score function penalizes the number of matches ($x_1$) by subtracting the percentage of mismatches ($x_6$), obtaining a lower value both when we have a low number of matches and when we have a high number of mismatches. To this value, a ratio is applied according to the number of possible matches ($x_5$), so that a comparison with a higher uncertainty obtains a worse score. Finally, another ratio is applied, again using the percentage of mismatches, to penalize comparisons with a high number of mismatches even more.

Looking at the final statistics obtained on the test dataset using this model (Table \ref{tab:MLStats}), and the CMC curve (Fig. \ref{cmcML}), we can clearly see that this proposal also improves the state-of-the-art result obtained by Adams and Aschheim's method.

\subsubsection{Multilayer perceptron}

For the last method proposed we obtain the results in the 5-fold CV in Table \ref{tab:foldsNN}:

\begin{table}[H]
\centering
\resizebox{0.49\textwidth}{!}{%
\begin{tabular}{@{}ccccc@{}}
\toprule
Fold                   & \multicolumn{2}{c}{Training} & \multicolumn{2}{c}{Validation} \\ \cmidrule(l){2-5}
                       & Average     & Maximum     & Average        & Maximum       \\ \midrule
\multicolumn{1}{c|}{1} & 0,0030      & 0,1159      & 0,0061         & 0,1429        \\
\multicolumn{1}{c|}{2} & 0,0030      & 0,1014      & 0,0105         & 0,3714        \\
\multicolumn{1}{c|}{3} & 0,0022      & 0,0725      & 0,0155         & 0,6286        \\
\multicolumn{1}{c|}{4} & 0,0034      & 0,0791      & \textbf{0,0004}         & \textbf{0,0294}        \\
\multicolumn{1}{c|}{5} & 0,0039      & 0,0935      & 0,0013         & \textbf{0,0294}        \\ \midrule
Average of folds       & 0,0031      & 0,0925      & 0,0068         & 0,2403        \\ \bottomrule
\end{tabular}
}
\caption{Normalized results of the average and maximum ranking obtained using a genetic algorithm to train a multilayer perceptron-based aggregation mechanism. In bold the best validation results.}
\label{tab:foldsNN}
\end{table}


For this method we cannot perform an in-depth analysis of how the score is obtained. We also achieve really good results for both training and validation sets, obtaining the best results with the fourth fold. Using the model of the fourth fold as our final model, in the test dataset we very promising results (Table \ref{tab:MLStats} and Fig. \ref{cmcML}).

\subsection{Comparison of methods}

Once we have seen how each proposal performs in relation to the Adams and Aschheim's method, we go on to compare and discuss the different proposals among them. We discussed their performance not only from a numerical point of view, but also taking into account the interpretability of the aggregation mechanism designed in each case.

\begin{table}[H]
\centering
\resizebox{0.49\textwidth}{!}{%
\begin{tabular}{@{}lccllllc@{}}
\toprule
                                                 & \multicolumn{1}{l}{Average} & \multicolumn{1}{l}{Q1} & Q2 & Q3 & P95 & P99 & \multicolumn{1}{l}{Max} \\ \midrule
\multicolumn{1}{l|}{AA} & 3.91                        & 1                      & 1  & 1  & 30  & 34  & 36                      \\
\multicolumn{1}{l|}{LO}                        & 2.14                        & 1                      & 1  & 1  & 6   & 16  & 34                      \\
\multicolumn{1}{l|}{OWA-Linear}                        & 2.86                        & 1                      & 1  & 1  & 11   & 22  & 36                      \\
\multicolumn{1}{l|}{OWHM-Linear}                        & 6.02                        & 1                      & 2  & 5  & 28   & 36  & 38                      \\
\multicolumn{1}{l|}{\begin{tabular}[c]{@{}l@{}}Choquet integral:\\ \hspace{0.4cm} $\mu$: $\lambda$\end{tabular}}                        & 3.14                        & 1                      & 1  & 1  & 16   & 33  & 35                      \\
\multicolumn{1}{l|}{\begin{tabular}[c]{@{}l@{}}Sugeno integral:\\ \hspace{0.4cm} $\mu$: $\lambda$\end{tabular}}                        & 5.28                        & 1                      & 3  & 5  & 18   & 25  & 35                      \\
\multicolumn{1}{l|}{LR}                        & 2.21                        & 1                      & 1  & 1  & 6   & 23  & \textbf{24}                      \\
\multicolumn{1}{l|}{SR}                        & 2.02                        & 1                      & 1  & 1  & 4   & 23  & 30                      \\
\multicolumn{1}{l|}{MLP}                        & \textbf{1.94}                        & 1                      & 1  & 1  & 4   & 18  & 28                      \\ \bottomrule
\end{tabular}
}
\caption{Comparison of ranking performance between all the proposals.}
\label{tab:allStats}
\end{table}

As can be seen in Table \ref{tab:allStats}, the machine learning methods achieve better results overall. Going into the details, we can see how the multilayer perceptron obtains the best average ranking with a slight improvement over symbolic regression. One interesting detail is that the linear regression model is the one with the lower maximum, which means that using this method would be the best way to find the positive comparison for the most difficult case in our dataset, although on average more comparisons would be needed for the other cases.

Regarding the fuzzy aggregation operators, they have performed worse than expected. While improving the Adams and Aschheim's method in the case of the OWA and the Choquet integral, these results are clearly behind the other two proposals. For this reason to compare the methods we discussed first the fuzzy aggregators, and then the rest of the machine learning models and our LO aggregation proposal.

\begin{figure}[H]
\centering
\includegraphics[width=3.5in]{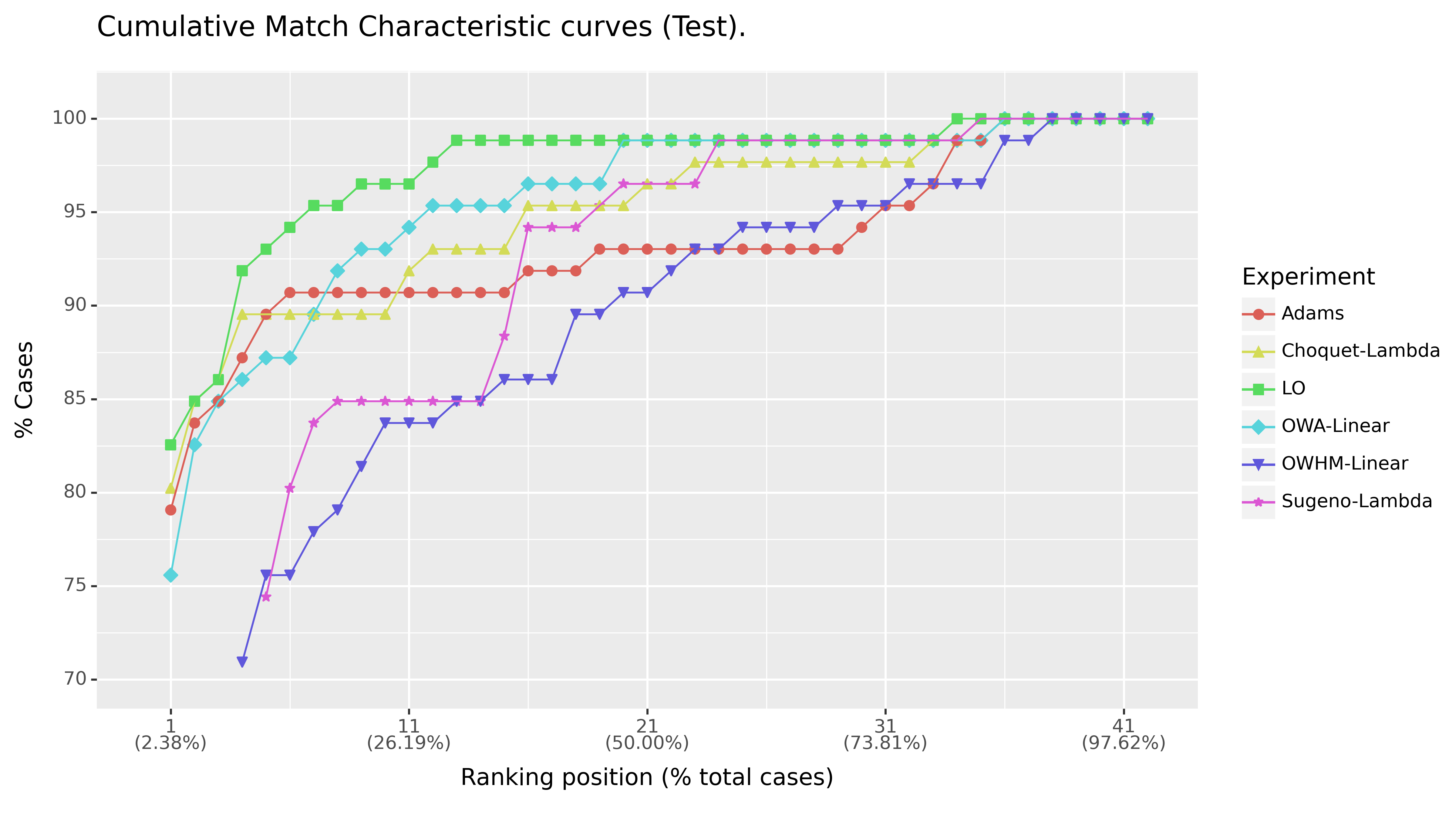}
\caption{CMC plot comparing the results of the best fuzzy aggregators and Adams and Aschheim's proposal in the test dataset. }
\label{cmcFuzzy}
\end{figure}

In Figure \ref{cmcFuzzy} we can see how for the fuzzy aggregators we can find three different behaviors. The OWHM performed much worse than the rest of the aggregators, the Sugeno integral performs much worse when trying to cover the first 90\% of the cases, but after that they slightly improve the results. Finally, the OWA with the linear weights and the Choquet integral are able to have a similar performance to the Adams and Aschheim's proposal at the first positions, improving the ranking after the first 15\% of cases, obtaining better results overall.

Looking at the rest of the methods, we can clearly see that they outperform the Adams and Aschheim's proposal:

\begin{figure}[H]
\centering
\includegraphics[width=3.5in]{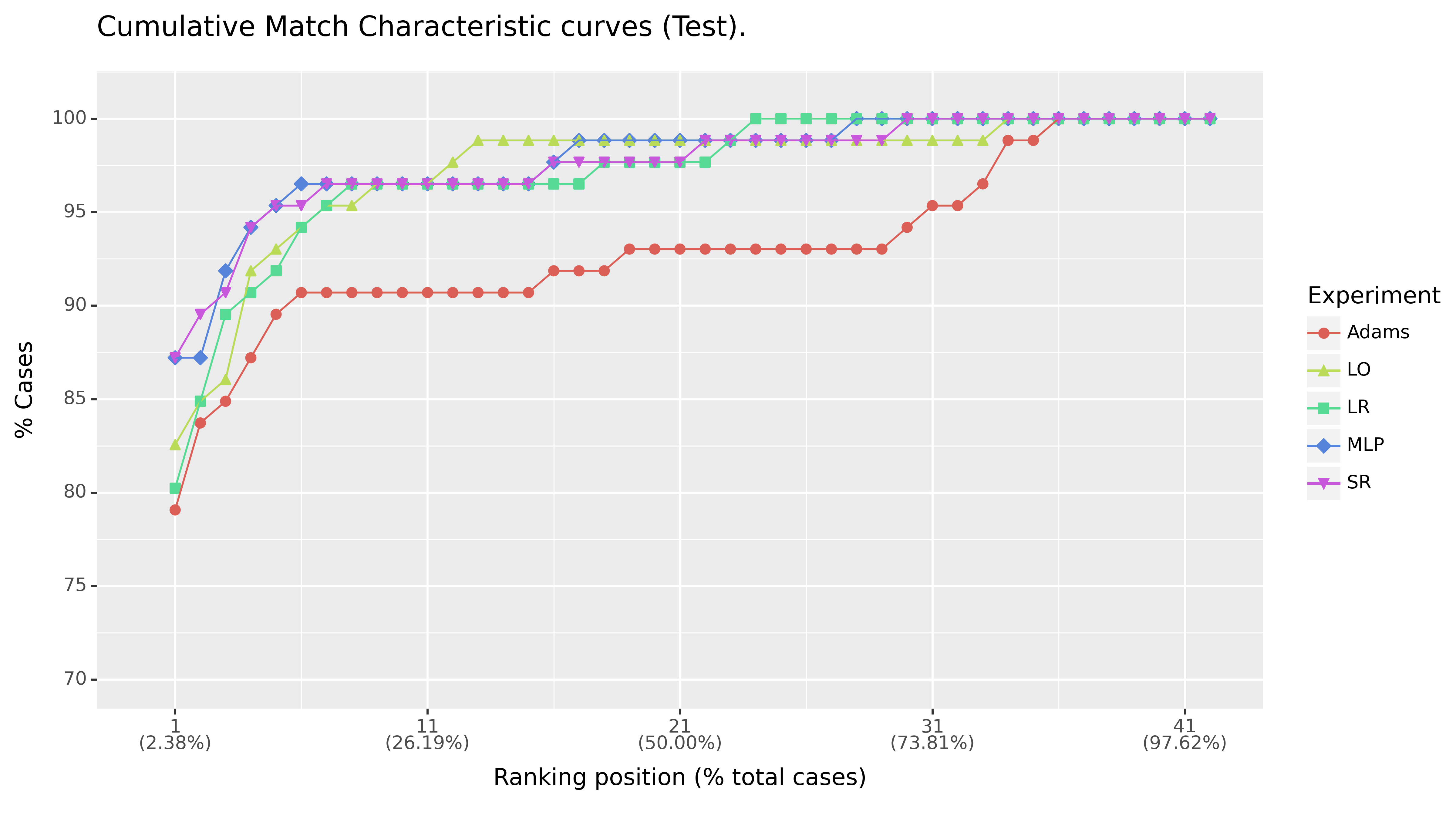}
\caption{CMC plot comparing the results of the machine learning aggregators, our LO aggregation mechanism proposal and the Adams and Aschheim's proposal in the test dataset. }
\label{cmcAll}
\end{figure}

All these methods have a similar behavior with small but interesting differences. Regarding the reordering of the criteria when using the LO, we can see how this small change improved greatly the percentage of cases covered achieving a result similar to the linear regression model. Both the linear regression and the reordering of the criteria proposals behaves very similarly to the Adams and Aschheim's method in the first positions of the ranking. However, with these methods it is possible to find 95\% of the correct AM-PM comparison with querying only the first 15\% of the comparisons, whereas Adams and Aschheim's proposal needs querying more than 70\% of the total comparisons to find the 95\% of correct AM-PM comparisons. Symbolic regression and multilayer perceptron are able to improve the remaining proposals by covering almost the 90\% of the correct AM-PM comparisons with the first position of the ranking.

Beyond the results obtained in terms of metrics, we can see a clear advantage in the proposals of symbolic regression, linear regression and our LO aggregator with respect to the multilayer perceptron. The simplicity of the equations obtained by both linear regression and symbolic regression makes these solutions easy for experts to interpret, unlike the multilayer perceptron proposal.

\section{Discussion and conclusions} \label{discussion}
\noindent

In this work, we studied the design of different aggregation mechanisms to generate a scoring system aimed at improving current methods for odontogram comparison in forensic scenarios. Specifically, to enhance ranking generation, we proposed and validated three different proposals. First, we examined and applied a variation of the state-of-the-art Adams and Aschheim system \cite{adams_computerized_2016}. Then, we also studied the feasibility of applying different well-known fuzzy aggregation operators, as the OWA, the OWHM, the Choquet integral and the Sugeno integral. Finally, we tried a machine learning-based approach using linear regression, symbolic regression and a multilayer perceptron as aggregation mechanisms, trained with real case scenarios to learn how to aggregate the criteria from the comparison of two odontograms. With these experiments we demonstrated the ability of interpretable machine learning methods to improve the performance of current methods for identifying deceased persons in DVI scenarios, while maintaining explainability and interpretability. This allows experts to understand how these methods compute the score, a critical step in this type of application.

With the first proposal, a simpler data-driven approach, we were able to significantly improve the results. With the second proposal we demonstrated the importance of making a good adjustment of the fuzzy aggregation methods in order to make them work in this particular application. Finally, our last proposal shows how machine learning models are able to extract underlying patterns on the data in order to extract new knowledge. Specially the symbolic regression, since thanks to its inherent variable selection and flexibility, outperformed the others approaches. On the contrary, the multilayer perceptron, although obtaining better results than symbolic regression, have shown that this slight advantage does not overcome the problem of obtaining a method that the expert can understand and gain knowledge from.

One of the main limitations of this work is the number of samples considered. Due to the sensitive nature of the information, it is difficult to obtain large and diverse databases to perform experiments with. Another limitation with respect to the data used is the existence of a low population diversity, since the cases used come only from two populations.

For future work, in addition to seeking more data from more varied populations, we also consider extending the proposal to different coding systems in addition to SCS. We also believe that the combination of different coding systems with different levels of granularity may be positive to deal with the tradeoff between simplicity and discriminative power. Finally, another future work is a study on the extraction of criteria when comparing odontograms, since a match, possible match or mismatch may have different relevance depending on whether the change in the dentition is more or less common.

\section*{Acknowledgments}
\noindent

This work was supported by grant CONFIA2 (PID2024-156434NB-I00) funded by the Spanish Ministry of Science, Innovation, and Universities and by 'EIC Accelerator - Seal of Excellence' project (09/942572.9/23), funded by Conserjer\'ia de Educaci\'on y Universidades de la Comunidad de Madrid.

Dr. Ib\'a\~nez’s work is funded by the Spanish Ministry of Science, Innovation and Universities under grant RYC2020-029454-I and by Xunta de Galicia by grant ED431F 2022/21. We wish to acknowledge the support received from the Centro de Investigaci\'on de Galicia "CITIC", funded by Xunta de Galicia and the European Union (European Regional Development Fund- Galicia 2014-2020 Program), by grant ED431G 2019/01.

This research project was made possible through the access granted by the Galician Supercomputing Center (CESGA) to its supercomputing infrastructure. The supercomputer FinisTerrae III and its permanent data storage system have been funded by the NextGeneration EU 2021 Recovery, Transformation and Resilience Plan, ICT2021-006904, and also from the Pluriregional Operational Programme of Spain 2014-2020 of the European Regional Development Fund (ERDF), ICTS-2019-02-CESGA-3, and from the State Programme for the Promotion of Scientific and Technical Research of Excellence of the State Plan for Scientific and Technical Research and Innovation 2013-2016 State subprogramme for scientific and technical infrastructures and equipment of ERDF, CESG15-DE-3114

\printbibliography

\newpage

\vfill

\end{document}